%% file: main.tex
\documentclass[letterpaper, 10pt, conference]{ieeeconf}
\IEEEoverridecommandlockouts  

\usepackage[utf8]{inputenc}
\usepackage{amsmath, amssymb, amsfonts}
\usepackage{graphicx}
\usepackage{cite}
\usepackage{url}
\usepackage[hidelinks]{hyperref}
\usepackage{booktabs}
\usepackage{float}
\usepackage{xcolor}
\usepackage{listings}
\lstdefinelanguage{json}{
  basicstyle=\ttfamily,
  numbers=left,
  numberstyle=\tiny, 
  stepnumber=1,
  numbersep=8pt,
  showstringspaces=false,
  breaklines=true,
  frame=lines,
  backgroundcolor=\color{white},
  morestring=[b]",
  moredelim=[s][\bfseries]{:}{\ },
  moredelim=[l][\bfseries]{,}
}

\usepackage[linesnumbered,ruled,vlined]{algorithm2e}
\usepackage{algorithmic}

\usepackage{subcaption}

\begin{document}

\title{

Scene-Aware Conversational ADAS with Generative AI for Real-Time Driver Assistance

}


\author{
Kyungtae Han$^{1}$, 
Yitao Chen$^{1}$, 
Rohit Gupta$^{1}$, 
Onur Altintas$^{1}$%
\thanks{$^{1}$InfoTech Labs, Toyota Motor North America, Mountain View, CA 94043, USA. 
Email: {\tt\small \{kt.han, yitao.chen, rohit.gupta, onur.altintas\}@toyota.com}}
}

\maketitle

\input{sections/abstract}

\input{sections/introduction}

\input{sections/related_work}

\input{sections/framework}

\input{sections/conversational_flow}

\input{sections/deployment}

\input{sections/evaluation}

\input{sections/results}

\input{sections/future_work}

\input{sections/conclusion}

\input{sections/acknowledge}

\bibliographystyle{IEEEtran}
\bibliography{references}
\end{document}

%% file: sections/abstract.tex
\begin{abstract}

While autonomous driving technologies continue to advance, current Advanced Driver Assistance Systems (ADAS) remain limited in their ability to interpret scene context or engage with drivers through natural language. These systems typically rely on predefined logic and lack support for dialogue-based interaction, making them inflexible in dynamic environments or when adapting to driver intent. This paper presents Scene-Aware Conversational ADAS (SC-ADAS), a modular framework that integrates Generative AI components including large language models, vision-to-text interpretation, and structured function calling to enable real-time, interpretable, and adaptive driver assistance. SC-ADAS supports multi-turn dialogue grounded in visual and sensor context, allowing natural language recommendations and driver-confirmed ADAS control. Implemented in the CARLA simulator with cloud-based Generative AI, the system executes confirmed user intents as structured ADAS commands without requiring model fine-tuning. We evaluate SC-ADAS across scene-aware, conversational, and revisited multi-turn interactions, highlighting trade-offs such as increased latency from vision-based context retrieval and token growth from accumulated dialogue history. These results demonstrate the feasibility of combining conversational reasoning, scene perception, and modular ADAS control to support the next generation of intelligent driver assistance.

\end{abstract}




%% file: sections/introduction.tex
\section{Introduction}

Advanced Driver Assistance Systems (ADAS) have become integral to modern vehicles, enhancing safety and driving comfort through features such as adaptive cruise control, lane keeping, and collision avoidance. However, current ADAS implementations are largely rule-based and reactive, offering limited flexibility in dynamic driving environments. They lack the ability to engage drivers through natural language conversation, interpret evolving road scenes, or adapt behavior in real time based on situational context and driver intent.

To enable more responsive and intuitive driver assistance, future systems must move beyond static, rule-based logic toward models that understand context and support natural interaction. As automation progresses, maintaining clear communication between the vehicle and driver is essential for safety and trust. Drivers must stay informed and engaged, confident in the system’s behavior. Intelligent interfaces that interpret natural language, reason over environmental cues, and respond conversationally can bridge the gap between human intent and machine action. Achieving this requires a unified framework combining language understanding, real-time scene perception, and structured control.

While prior works have explored components of this challenge, key limitations remain. Personalized ADAS systems~\cite{liao2023digitaltwin, liao2024personalizationreview} emphasize behavioral modeling but lack interactive and scene-aware capabilities. 
Current in-vehicle voice assistants primarily support basic command interfaces related to navigation, media, and settings, without multi-turn dialogue or deep integration with ADAS functionalities.
Tool-augmented large language models (LLMs), such as ReAct~\cite{yao2023react} and ToolLLM~\cite{qin2024toolllm}, enable structured function calling but have not yet been deployed in driving systems. Vision-language models (VLMs), including BLIP-2~\cite{li2023blip} and GPT-4V~\cite{openai2023gpt4}, extract semantic scene information but are not used in closed-loop interactive pipelines.
Recent efforts such as LaMPilot~\cite{ma2024lampilot} and \textit{Drive As You Speak}~\cite{cui2024drive} introduce LLM-driven decision-making for driving tasks, but do not fully address the demands of scene-aware, conversational ADAS or evaluate real-time modular system behavior.

\begin{figure}[t]
    \centering
    \includegraphics[width=0.50\textwidth, trim=30pt 0pt 0pt 0pt, clip]{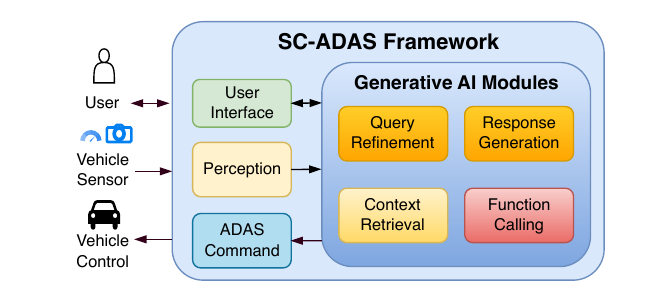}
    \caption{
        Overview of SC-ADAS. The system combines four Generative AI modules with user interface, sensors, and control for real-time, scene-aware conversational interaction.}
    \label{fig:scadas_overview}
\end{figure}

This paper presents Scene-Aware Conversational ADAS (SC-ADAS), a modular framework that integrates LLM-based reasoning, vision-to-text scene understanding, and structured function calling into a closed-loop driver assistance system.
As illustrated in Fig.~\ref{fig:scadas_overview}, the system consists of four Generative AI-powered modules, integrated with the user interface, vehicle sensors, and control components to support real-time, scene-aware conversational interaction. SC-ADAS supports multi-turn dialogue, allowing drivers to issue natural language queries, receive context-aware recommendations, and confirm executable ADAS commands through structured function invocation.

This work addresses the following research question: \textit{Can Generative AI enable real-time, scene-aware, multi-turn driver assistance that is both interpretable and efficient enough for practical ADAS deployment?} To investigate this, we develop SC-ADAS and evaluate it in a simulated driving environment~\cite{dosovitskiy2017carla}, analyzing end-to-end latency, token usage growth, and system behavior across different modular configurations.

The main contributions of this work are:
\begin{itemize}
\item Introduction of SC-ADAS, a modular framework integrating LLMs, vision-to-text models, and structured ADAS control;
\item Design of a multi-turn conversational pipeline grounded in real-time visual context and driver intent;
\item Empirical evaluation of latency and token usage across modular configurations, highlighting design trade-offs;
\item Identification of deployment strategies for GenAI-powered ADAS, including latency reduction and memory optimization.
\end{itemize}


%% file: sections/related_work.tex
\section{Related Work}
\label{sec:related}

\subsection{ADAS Systems and Personalization}

Modern ADAS systems provide features such as adaptive cruise control, lane keeping, and collision avoidance, but remain largely rule-based and reactive, offering limited flexibility in dynamic driving environments. Recent research has explored personalization in driver behavior modeling~\cite{liao2023digitaltwin, liao2024personalizationreview}, including online prediction of lane change preferences and comprehensive reviews of personalized driving datasets and modeling approaches. However, these efforts primarily target trajectory-level or behavioral adaptation, and do not support real-time, scene-aware, conversational systems.

\subsection{Voice Assistants in Vehicles}
Early systems such as Adasa~\cite{lin2018adasa} demonstrated the potential for conversational interfaces to access ADAS features. However, most commercial in-vehicle assistants remain limited to basic command-driven interactions, focusing on navigation, media control, and system settings without support for multi-turn dialogue, contextual understanding, or real-time integration with ADAS functionalities~\cite{du2024towards}.

\subsection{Tool-Use LLMs and Function Calling}

Recent advances in LLMs have enabled reasoning-driven tool use through prompting techniques such as ReAct~\cite{yao2023react} and ToolLLM~\cite{qin2024toolllm}. These models can interpret user queries, invoke external functions, and generate structured outputs. However, they have not been applied in real-time driving contexts and do not integrate with vehicle sensors or control systems.

\subsection{Vision-Language Models for Scene Understanding}

VLMs, including BLIP-2~\cite{li2023blip} and GPT-4V~\cite{openai2023gpt4}, have demonstrated strong performance in extracting semantic understanding from images and multimodal scenes. Recent surveys~\cite{zhou2024vlmadas, cui2024multimodal} have reviewed the potential of applying VLMs to autonomous driving, highlighting opportunities in perception, planning, and driver interaction. However, current VLMs are typically used for passive perception tasks rather than being integrated into closed-loop, real-time interaction systems for ADAS operation. Their outputs remain disconnected from active driver engagement or immediate vehicle control.

\subsection{LLM-Based Driving Agents and Instruction Following}

Several recent studies have proposed leveraging LLMs to interpret natural language commands in autonomous driving scenarios. LaMPilot~\cite{ma2024lampilot} formulates driving as a language-program generation task, enabling symbolic action code creation in simulation environments. Personalized Autonomous Driving with LLMs~\cite{cui2024personalized} presents field experiments where large language models are used to personalize driving behavior based on natural language instructions. The \textit{Drive As You Speak} framework~\cite{cui2024drive} conceptualizes LLMs as decision-making cores within interactive driving architectures. 

While prior research has explored personalization, voice-based interaction, and language-enabled reasoning in autonomous driving, these efforts often lack real-time modular service composition and closed-loop scene-aware interaction. In contrast, SC-ADAS uniquely integrates multi-turn dialogue, scene-grounded context understanding, structured ADAS actuation, and modular service evaluation, representing a practical step toward deploying Generative AI-powered ADAS systems in real-world environments.

%% file: sections/framework.tex
\section{SC-ADAS Framework}
\label{sec:framework}

\subsection{Design Objectives}

The SC-ADAS framework bridges the gap between natural driver input and real-time ADAS behavior by integrating perception, reasoning, and actuation in a modular conversational loop. The design objectives are as follows:

\begin{enumerate}
    \item \textbf{Multimodal Context Understanding:}
    The system processes a driver’s voice command as the primary query input and supplements it with multimodal context, including real-time sensor readings and vision-derived scene descriptions. The complete driving context is defined as:

    \begin{equation}
        C = \{C_{\text{sensing}}, C_{\text{document}}\}
    \end{equation}
    where $C_{\text{sensing}}$ includes real-time sensor data (e.g., speed, camera images), and $C_{\text{document}}$ refers to static knowledge such as maps and prior preferences.

    \item \textbf{Context-Aware Command Generation:}  
    Given a refined query \( Q_{\text{refined}} \), context \( C \), and dialogue history \( H \), the system generates an ADAS command set \( S \) through a learned mapping:
    \begin{equation}
        S = f(Q_{\text{refined}}, C, H)
    \end{equation}
    ensuring commands align with both user intent and situational constraints.

    \item \textbf{Low-Latency Processing:}  
    All processing stages are required to complete within real-time constraints to maintain safe and timely driver assistance.


\end{enumerate}

\subsection{System Architecture Overview}

As illustrated in Fig.~\ref{fig:scadas_overview}, the SC-ADAS system is organized as a modular pipeline composed of four Generative AI-powered components. Spoken driver queries are refined, augmented with environmental context, processed through dialogue generation, and ultimately translated into structured ADAS commands. Each module in the architecture performs a distinct role in the conversational loop. Their functions are described below.

\subsection{Modular Pipeline Components}

\begin{table}[t]
\centering
\caption{Summary of SC-ADAS Modular Components}
\label{tab:module_summary}
\begin{tabular}{|p{1.3cm}|p{6.5cm}|}
\hline
\textbf{Module} & \textbf{Function} \\
\hline
Query\newline Refinement & Interprets the driver’s voice command and adds metadata such as whether context or actuation is needed. \\
\hline
Context\newline Retrieval & Collects relevant multimodal information, including real-time sensor data and vision-to-text scene descriptions. \\
\hline
Response\newline Generation & Crafts a natural language reply based on query intent, retrieved context, and dialogue history. \\
\hline
Command\newline Generation & Translates confirmed user intent into structured ADAS commands using function calling. \\
\hline
\end{tabular}
\end{table}

Table~\ref{tab:module_summary} summarizes the roles of the four primary modules.

\textbf{1) Query Refinement Module:}  
This module interprets the driver’s voice command, which is converted to text via speech recognition, and produces a refined query \( Q_{\text{refined}} \) and metadata \( M \). Metadata flags indicate whether context retrieval or actuation is required. Formally:
\begin{equation}
(Q_{\text{refined}}, M) = f_{\text{refine}}(Q_{\text{user}}, H)
\end{equation}

\textbf{2) Context Retrieval Module:}  
Triggered when additional context is needed, this module gathers sensing and document data:
\begin{equation}
C = f_{\text{retrieval}}(Q_{\text{refined}}, M)
\end{equation}
Vision-to-text models are used to interpret camera inputs into natural language scene descriptions.

\textbf{3) Response Generation Module:}  
Given the refined query, context, prior action set, and dialogue history, this module generates a natural language reply \( R \):
\begin{equation}
R = f_{\text{response}}(Q_{\text{refined}}, C, S, H)
\end{equation}
This stage supports recommendation, clarification, and confirmation interactions.

\textbf{4) Command Generation Module:}  
Upon driver confirmation, this module formats intent into structured ADAS commands, compatible with vehicle or simulator control APIs. For example:

\begin{lstlisting}[language=json,caption={Example function call output.},label={lst:function_call_result}]
{ "name": "set_speed",
  "arguments": { "speed": 50 }}
\end{lstlisting}

The structured output is forwarded to the ADAS backend for execution.

%% file: sections/conversational_flow.tex
\section{Conversational Flows}
\label{sec:conversational_flows}

\subsection{Overview of Interaction Patterns}

SC-ADAS supports two primary types of conversational interaction: \textit{Informational} and \textit{Actionable}. Both flows leverage the modular reasoning and response pipeline described earlier, adapting processing paths based on user intent.

    

\begin{figure}[t]
    \centering
    \begin{subfigure}[t]{0.50\textwidth}
        \centering
        \includegraphics[width=\linewidth, trim=10pt 0pt 10pt 0pt, clip]{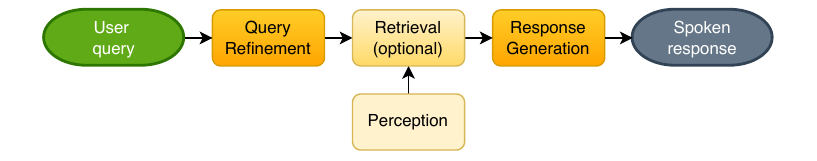}
        \caption{Informational interaction: The system interprets the query, optionally retrieves context, and responds with a recommendation.}
        \label{fig:informational_flow}
    \end{subfigure}

    \vspace{10pt}

    \begin{subfigure}[t]{0.50\textwidth}
        \centering
        \includegraphics[width=\linewidth, trim=10pt 0pt 10pt 0pt, clip]{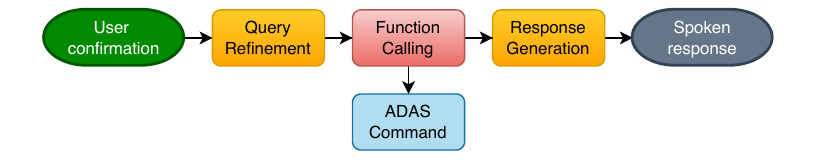}
        \caption{Actionable interaction: The system processes user confirmation, generates a structured command, and executes it with acknowledgment.}
        \label{fig:actional_flow}
    \end{subfigure}
    \caption{Interaction flows supported by SC-ADAS. (a) handles general recommendations and status queries; (b) handles confirmed actions resulting in ADAS control changes.}
    \label{fig:interaction_flows}
\end{figure}

As illustrated in Fig.~\ref{fig:interaction_flows}:

\textbf{(a) Informational Interaction:}  
The driver issues a query requesting a recommendation or status update. The system refines the query, optionally retrieves real-time context (e.g., current speed, scene description), and generates a natural language response.

\textbf{(b) Actionable Interaction:}  
The driver confirms an action that alters ADAS behavior. The system refines the confirmed intent, generates a structured command via function calling, and executes the command after validation.

This modular routing enables SC-ADAS to dynamically adjust its reasoning and control steps while ensuring that all vehicle-level actuation is explicitly confirmed before execution.

\subsection{Example of Multi-Turn Conversational Flow}

To illustrate SC-ADAS operation during runtime, we present a two-turn scenario involving both interaction types. Each turn selectively activates different modules based on the query’s metadata and system context.

\subsubsection*{Turn 1: Informational Request}

\begin{enumerate}
\item \textbf{Driver Query:}  
“Could you recommend a safe speed for this road?”

\item \textbf{Query Refinement:}  
The system refines the input to:  
\( Q_{\text{refined}} = \) ``What speed is safe for this road?''  
with metadata \( M = \{\text{sensing\_required} = \text{true}\} \).

\item \textbf{Context Retrieval:}  
Retrieved context includes:  
\( C_{\text{vision}} = \) ``urban road, moderate traffic''  
and  
\( C_{\text{motion}} = \) ``current speed 42 MPH''.

\item \textbf{Response Generation:}  
Based on the retrieved context and dialogue history, the system responds:  
``Based on the detected urban road and moderate traffic conditions, I recommend setting the speed to 40 MPH. Would you like to apply this setting?''
\end{enumerate}

\vspace{0.2cm}

\subsubsection*{Turn 2: Command Confirmation (Actionable Interaction)}
\begin{enumerate}
\vspace{-0.4cm}
    \setcounter{enumi}{4}
    \item \textbf{Driver Query:}  
    ``Yes, go ahead.''

    \item \textbf{Query Refinement:}  
    Refined as:  
    \( Q_{\text{refined}} = \) ``Yes, set the speed to 40 MPH''  
    with metadata \( M = \{\text{actuation\_required} = \text{true}\} \).

    \item \textbf{Command Generation:}  
    Following the structure shown in Listing~\ref{lst:function_call_result}, the system generates a \texttt{set\_speed} command with the argument \texttt{speed = 40}.

    \item \textbf{Command Execution and Acknowledgment:}  
    The command is transmitted to the ADAS backend for execution, and the system acknowledges with:  
    ``Speed has been set to 40 MPH.''
\end{enumerate}

%% file: sections/deployment.tex
\section{System Deployment}
\label{sec:deployment}

\subsection{System-Level Deployment}

\begin{figure}[t]
  \centering
  \includegraphics[width=0.4\textwidth]{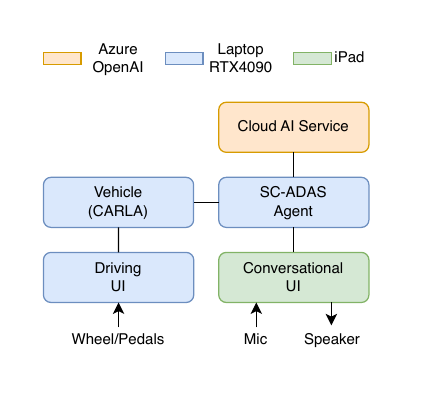}
  \vspace{-20pt}
  \caption{SC-ADAS system implementation architecture. The laptop runs the CARLA simulator, SC-ADAS agent, and UI server. The iPad serves as the conversational interface. GPT-4o is accessed remotely via Azure OpenAI using prompt-based interaction.}
  \label{fig:architecture}
\end{figure}

The system-level implementation architecture of SC-ADAS is illustrated in Fig.~\ref{fig:architecture}. The framework operates entirely on a high-performance laptop equipped with an NVIDIA RTX 4090 GPU and Intel Core i9 CPU, which hosts the CARLA simulator, vehicle control modules, and the SC-ADAS agent.

The driver interacts with SC-ADAS through an iPad running a browser-based conversational interface. The iPad captures voice queries via its built-in microphone, renders system responses using on-device text-to-speech, and displays the query-response history. Communication between the iPad and the backend agent is conducted over a local Wi-Fi network using WebSocket and HTTP protocols.

The SC-ADAS agent manages all four modular components described in Section~\ref{sec:framework}, namely query refinement, context retrieval, response generation, and command generation. Each module is implemented as a prompt-based reasoning service interacting with GPT-4o hosted on Azure OpenAI. Prompts are dynamically constructed based on driver queries, sensor context, and conversation history, enabling flexible orchestration of modular reasoning.

This architecture balances local perception and control with cloud-based reasoning, supporting real-time operation without requiring on-device model inference. The system achieves low-latency interaction while leveraging state-of-the-art Generative AI capabilities for scene-aware dialogue and adaptive ADAS command generation.

\subsection{Simulation Environment}

SC-ADAS is implemented and tested using the CARLA Simulator (v0.9.15)~\cite{dosovitskiy2017carla}, providing high-fidelity simulated sensor data, road scene diversity, and scripted driving behaviors. The ego vehicle is equipped with simulated telemetry, including real-time speed, and front-facing camera images, all streamed to the SC-ADAS backend for reasoning and actuation.


\subsection{User Interface Setup}

\begin{figure}[t]
    \centering
    \includegraphics[width=0.95\columnwidth]{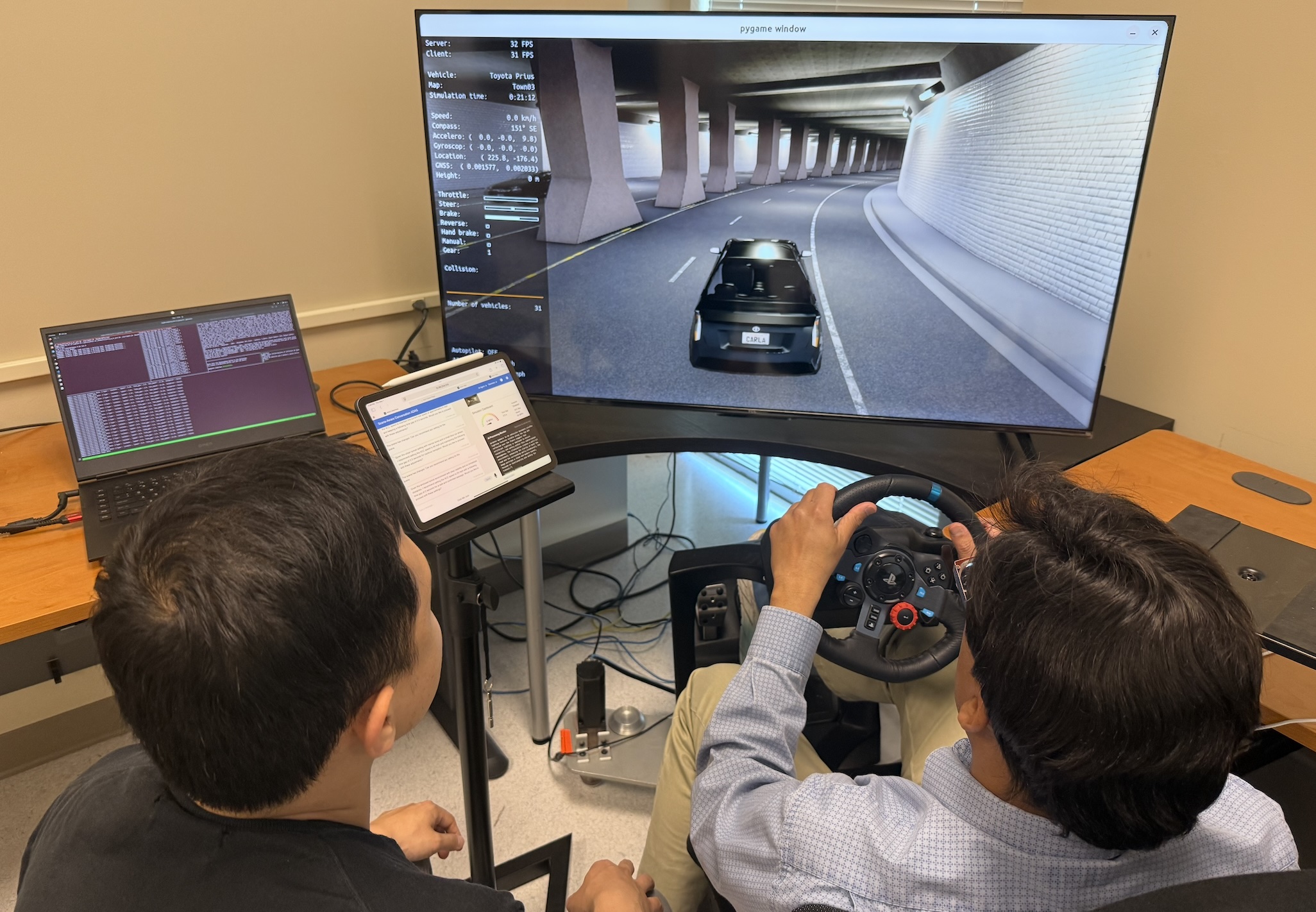}
    \caption{Testing setup using CARLA simulator and browser-based user interface on an iPad. The iPad handles voice input, speech output, and dialogue display.}
    \label{fig:test_setup}
\end{figure}

The complete experimental setup, including the iPad conversational interface and simulator, is shown in Fig.~\ref{fig:test_setup}.
The iPad interface supports the following functionalities:
\begin{itemize}
    \item \textbf{Speech Recognition:}  
    Converts spoken driver queries into text using the browser-based Web Speech API.
    
    \item \textbf{Text-to-Speech:}  
    Delivers SC-ADAS system responses as spoken messages using the built-in Web Speech Synthesis API.
    
    \item \textbf{Dialogue Display:}  
    Presents real-time system responses, context information, and status updates on the screen.
\end{itemize}

Communication between the iPad and the backend is maintained over a low-latency Wi-Fi connection using WebSocket for streaming telemetry updates and HTTP for cloud-based API transactions.

\subsection{Backend Processing}

The backend server orchestrates local data preparation and cloud-based reasoning as follows:

\begin{itemize}
    \item \textbf{Local Preprocessing Services}:  
    Local modules collect real-time telemetry data (e.g., ego speed, vehicle state, front-facing camera images) and process spoken driver input into text format. This raw information is organized into structured prompt templates, including the driver's query, live sensing data, and minimal dialogue history.

    \item \textbf{Cloud-Based Reasoning and Generation}:  
    All reasoning and decision-making modules, including query refinement, metadata extraction (e.g., sensing or actuation requirements), context retrieval, response generation, and command generation, are executed through a cloud-hosted Generative AI model. Specifically, GPT-4o, accessed via Azure OpenAI endpoints, performs natural language reasoning, environmental grounding, and structured function calling.
\end{itemize}

At each interaction turn, the backend server constructs a prompt that encapsulates the refined conversation history and relevant sensor data, then submits this prompt to the cloud model for processing. The model interprets the query, retrieves and fuses context if necessary, generates conversational responses, and constructs structured ADAS commands when confirmed.

To improve responsiveness, the maximum output token length is limited to 300 tokens for standard interactions and extended to 500 tokens when vision-derived contextual information is incorporated.
    
\subsection{System Integration and Communication}

The overall communication architecture is designed to support modular, real-time interaction between components:

\begin{itemize}
    \item ZeroMQ is used for telemetry data streaming from the CARLA server to the backend service.
    \item REST API calls manage cloud interactions for Generative AI services.
    \item WebSocket communication handles real-time synchronization between the backend and the iPad UI.
\end{itemize}

The modular communication strategy allows each system component (simulation, backend, reasoning model, and interface) to be deployed flexibly across different machines or scaled independently if needed.

%% file: sections/evaluation.tex
\section{Evaluation}
\label{sec:evaluation}

\subsection{Interaction Types Observed}

To evaluate SC-ADAS across different real-world usage scenarios, we ran a continuous interactive driving session and later categorized the observed interactions into three primary types:

\begin{itemize}
    \item \textbf{Conversational-Only Service}: Driver queries are processed into natural language responses without ADAS actuation or scene context retrieval.
    \item \textbf{Conversational ADAS Service}: Structured ADAS commands are issued upon driver confirmation, without using vision-derived context.
    \item \textbf{Scene-Aware Conversational Service}: 
    The system processes driver queries by retrieving real-time vision context, interpreting the scene, and generating a context-aware conversational response, without issuing ADAS control commands.
\end{itemize}

All interaction types share the Refiner and Responder modules, enabling isolated analysis of the Vision and Actuator module impacts.

\subsection{Evaluation Scenarios}

To comprehensively evaluate system performance, we conducted multiple interaction turns in a simulated driving session. These included scene-related queries (e.g., identifying objects in the environment), speed recommendation requests for adaptive cruise control (ACC), and confirmation commands for setting ADAS parameters. This mix of informational and actionable interactions ensures that all major system capabilities are exercised under realistic conditions.

SC-ADAS is evaluated along two dimensions: scene understanding effectiveness and real-time modular service efficiency.

\subsubsection{Scene-Based Speed Recommendation Evaluation}

To assess scene-aware reasoning, we design a $2\times3$ matrix combining two road types (highway in Town4 and downtown in Town3) and three weather conditions (clear, rainy, and foggy).  
In each case, the system receives the spoken query:  
\textit{``What speed do you recommend for this road?''}

\begin{figure*}[t]
    \centering

    \begin{minipage}[b]{0.3\textwidth}
      \centering
      \includegraphics[width=\linewidth]{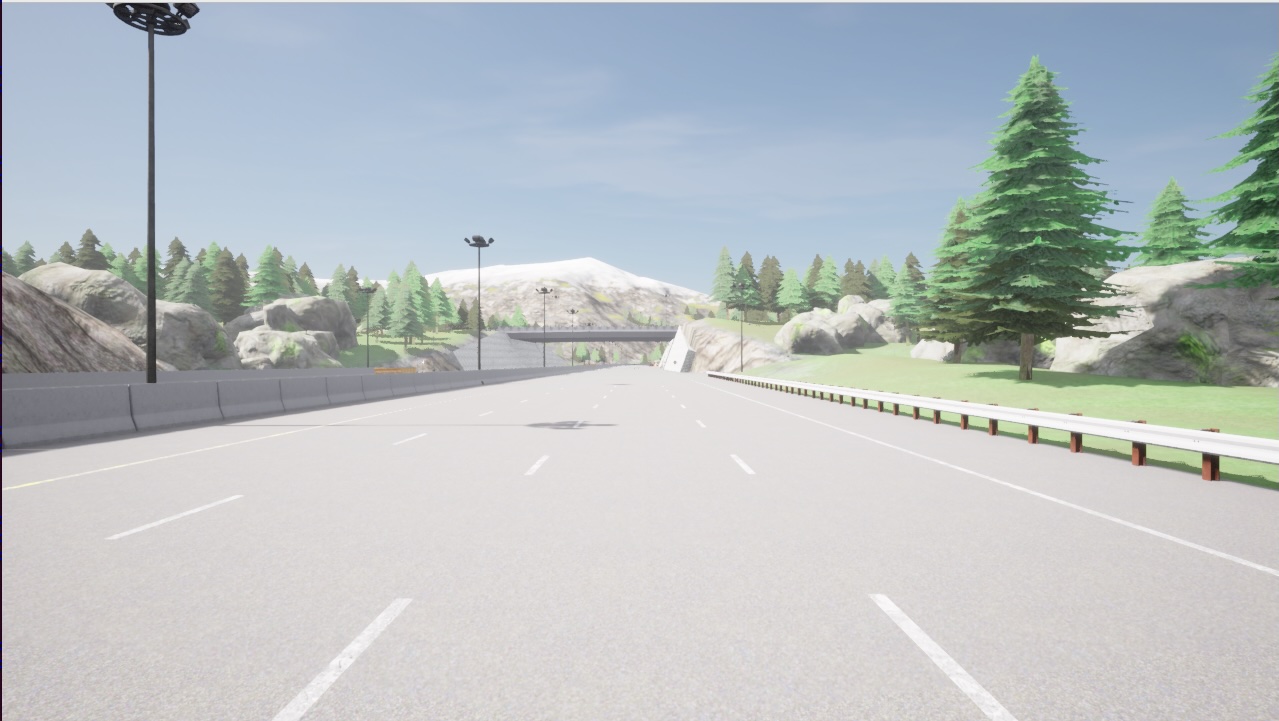}
    \end{minipage}
    \hspace{0.02\textwidth}
    \begin{minipage}[b]{0.3\textwidth}
      \centering
      \includegraphics[width=\linewidth]{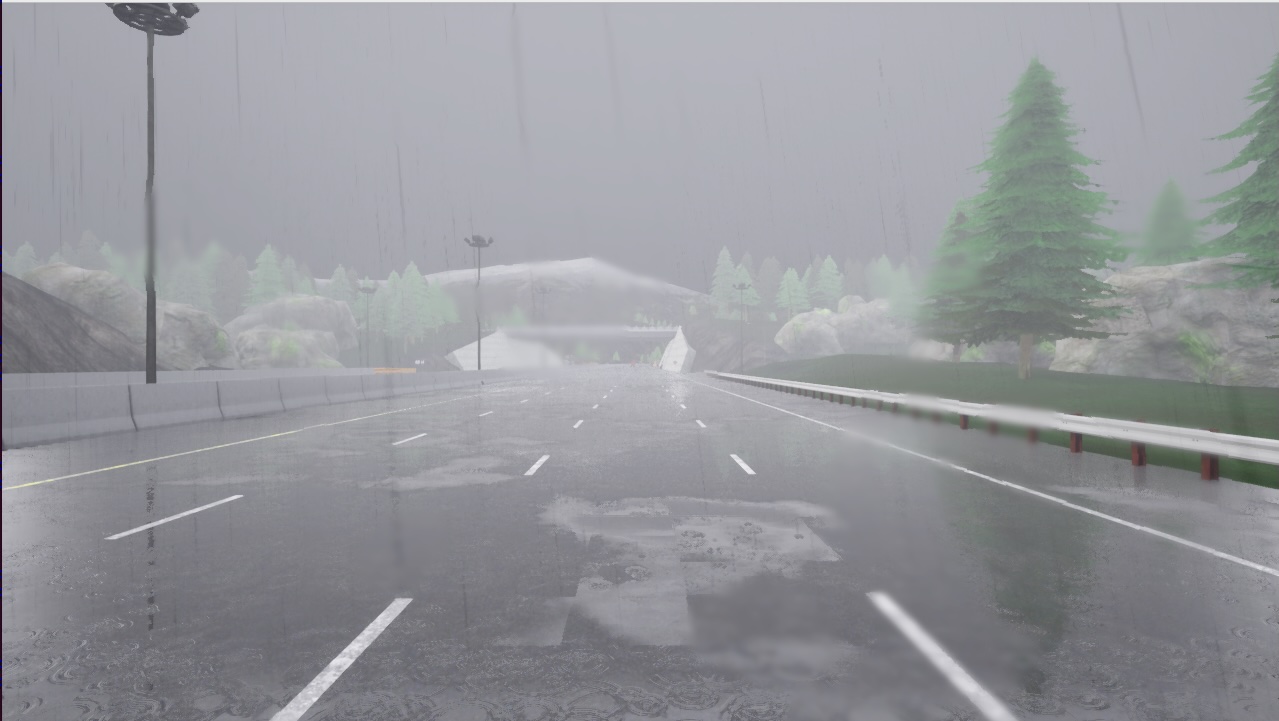}
    \end{minipage}
    \hspace{0.02\textwidth}
    \begin{minipage}[b]{0.3\textwidth}
      \centering
      \includegraphics[width=\linewidth]{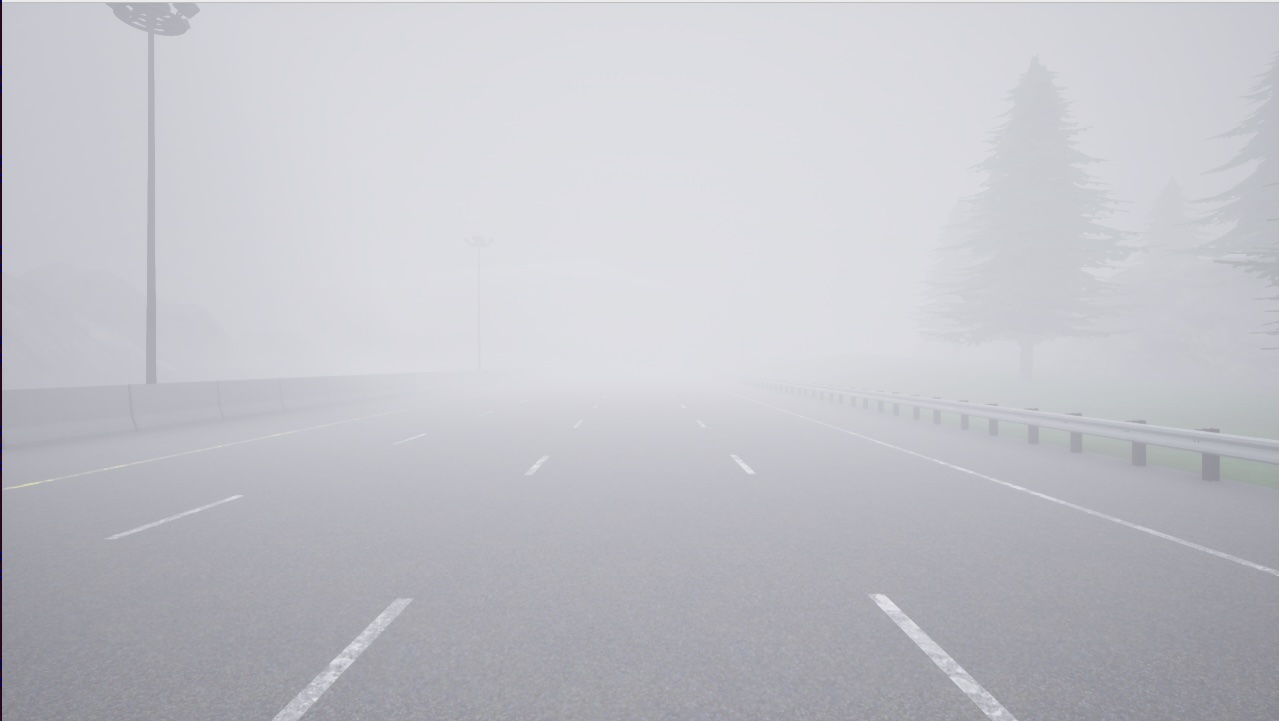}
    \end{minipage}

    \vspace{1em}

    \begin{minipage}[b]{0.3\textwidth}
      \centering
      \includegraphics[width=\linewidth]{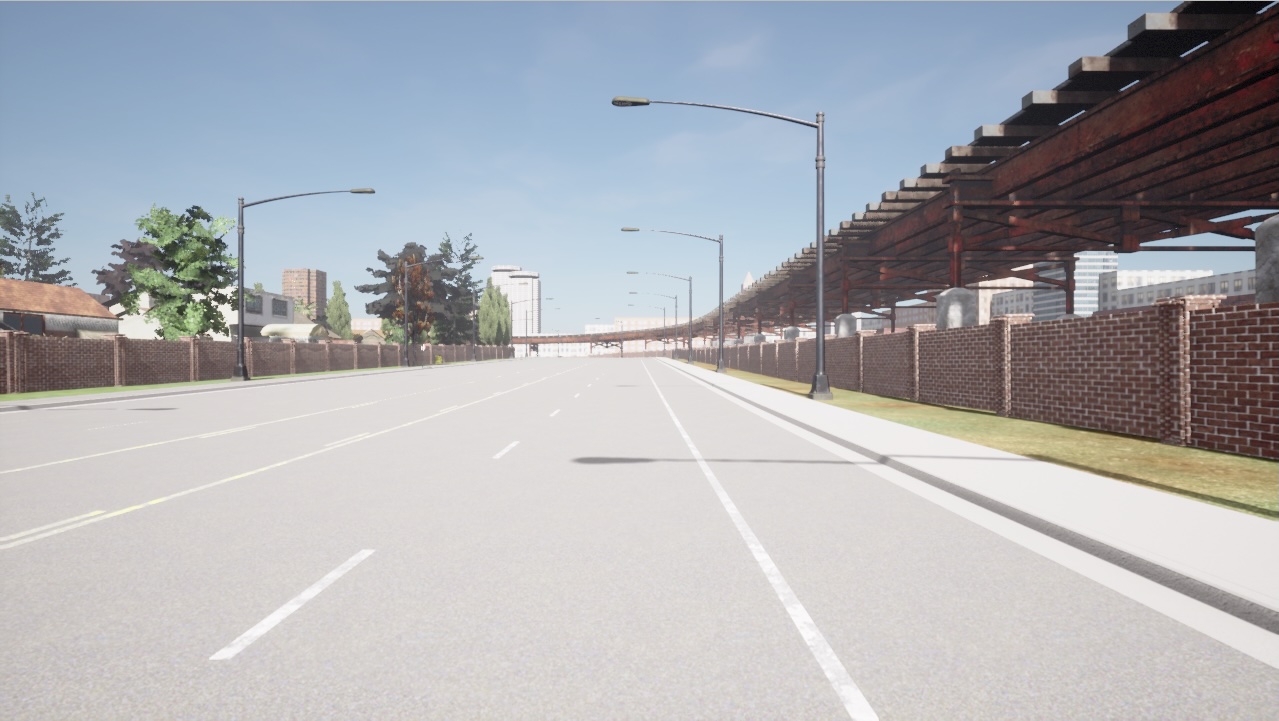}
    \end{minipage}
    \hspace{0.02\textwidth}
    \begin{minipage}[b]{0.3\textwidth}
      \centering
      \includegraphics[width=\linewidth]{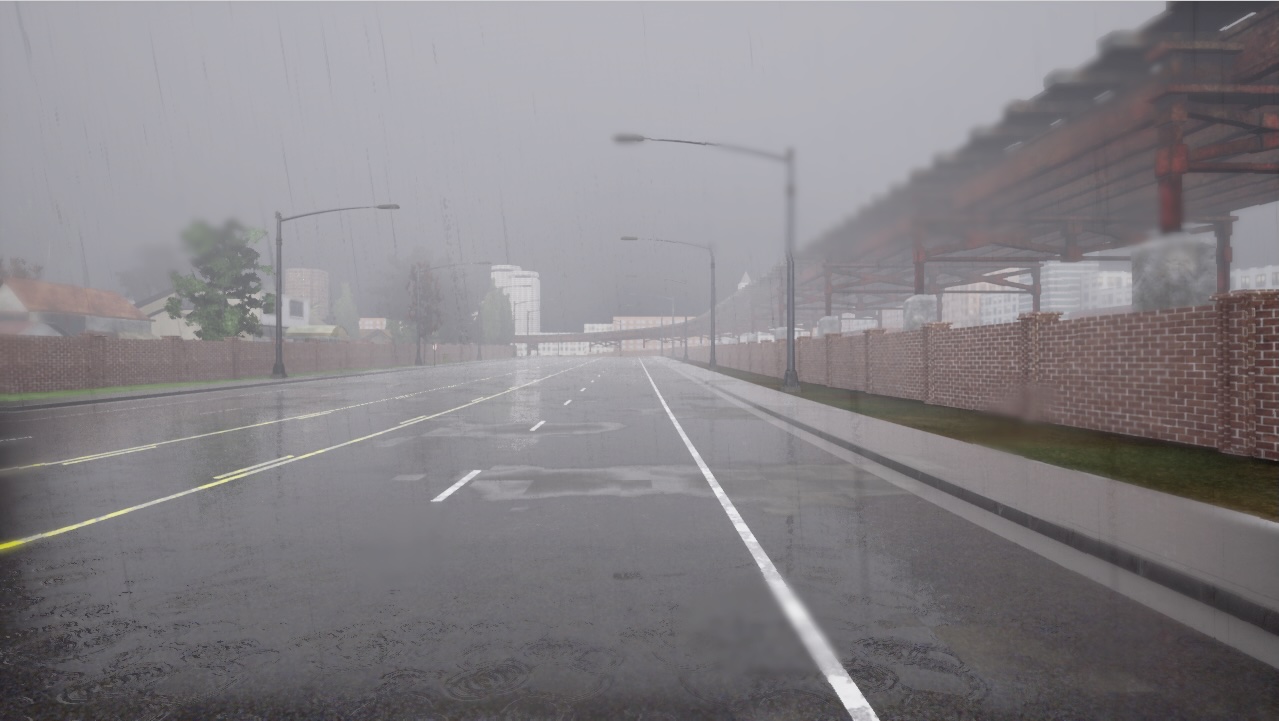}
    \end{minipage}
    \hspace{0.02\textwidth}
    \begin{minipage}[b]{0.3\textwidth}
      \centering
      \includegraphics[width=\linewidth]{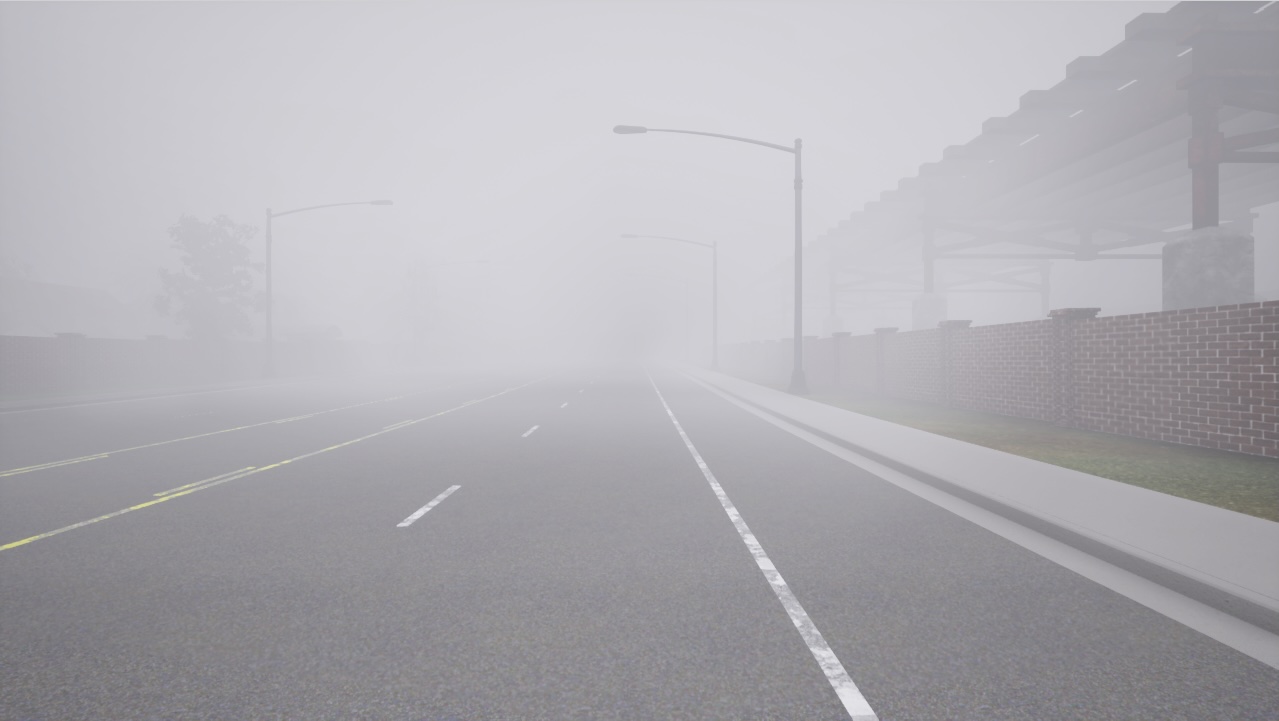}
    \end{minipage}
    \caption{Scene-based evaluation scenarios in CARLA Simulator: clear, rainy, and foggy conditions on highway (top row) and downtown (bottom row) roads. Each screenshot corresponds to a test instance where SC-ADAS provides speed recommendations.}
    \label{fig:weather_screenshots}
\end{figure*}

Front-camera images captured from the CARLA Simulator were used to simulate these conditions. A screenshot from each of the six test cases is shown in Fig.~\ref{fig:weather_screenshots}.

\subsubsection{Scene Consistency Evaluation}

To assess the consistency of SC-ADAS recommendations in repeated contexts, we designed an additional evaluation in which the ego vehicle returned to visually similar scenes during the same driving session. At each revisit, the driver issued the same adaptive cruise control query to observe whether the system would produce consistent responses.

This scenario helps test the stability of vision-based interpretation and downstream reasoning in a multi-turn dialogue setting, where prior user confirmations or preferences might reasonably influence future responses. Differences in output can reveal how variability in scene descriptions affects recommendation generation.

\subsubsection{Modular Service Performance Evaluation}

For real-time performance evaluation, we conduct full conversational service invocations across the two service types, including both informational (recommendations) and actionable (ADAS setting) interactions. This enables measurement of system latency and token usage evolution across different service complexities.

\subsection{Evaluation Metrics}

We define two primary evaluation metrics:

\begin{itemize}
    \item \textbf{Module Processing Latency}:  
    For each service invocation during simulated driving sessions, we measured the processing time of each reasoning module (Refiner, Vision, Actuator, Responder) as the duration between sending a prompt to the cloud-based Generative AI and receiving a response.  
    Front-end processing times, such as speech recognition and UI rendering, were excluded as their contribution was negligible.

    \item \textbf{Token Usage}:  
    Measured as input (uplink) and output (downlink) token counts per service invocation, tracked separately for each reasoning module.
\end{itemize}

These metrics capture system responsiveness, cloud service demands, and scalability characteristics.

%% file: sections/results.tex
\section{Results and Discussion}
\label{sec:results}

We evaluate SC-ADAS from two perspectives: its effectiveness in producing scene-aware driver assistance, and its efficiency in real-time operation under different modular service configurations.

\subsection{Effectiveness: Scene Understanding and Decision Quality}

\subsubsection{Scene-Based Speed Recommendation}

SC-ADAS produced differentiated speed recommendations across the six test scenarios shown in Fig.~\ref{fig:weather_screenshots}. Table~\ref{tab:scene_speed} summarizes the recommended speeds for each combination of road type and weather condition.

\begin{table}[h]
\centering
\caption{Recommended Speed (MPH) Across Road and Weather Conditions}
\label{tab:scene_speed}
\begin{tabular}{|c|c|c|c|}
\hline
\textbf{Road Type} & \textbf{Clear} & \textbf{Rainy} & \textbf{Foggy} \\
\hline
Highway (CARLA Town4)            & 65             & 40             & 30             \\
Downtown (CARLA Town3)           & 45             & 20             & 20             \\
\hline
\end{tabular}
\end{table}

The results demonstrate that SC-ADAS’s vision-to-text module successfully interprets environmental conditions, enabling adaptive and context-aware speed recommendations. The outputs reflect human-like sensitivity to road type and visibility, validating the system’s ability to ground language understanding in real-world scenes.

\subsubsection{Consistency Across Multi-Turn Scene Queries}

To evaluate the consistency of recommendations over time, we conducted additional tests where the driver returned to similar scenes within the same driving session and issued follow-up queries. While the system correctly interpreted the environment and produced relevant responses, we observed slight variations in the recommended ACC settings across turns, even when the vehicle revisited similar locations under comparable visual conditions.

This variation is likely due to the generative nature of the vision-to-text module, where subtle differences in textual scene descriptions influenced downstream reasoning. Although each description was semantically valid, wording variability introduced minor fluctuations in the resulting recommendations.

This observation highlights a trade-off between natural language flexibility and behavioral consistency. To improve stability, future iterations of SC-ADAS may incorporate more structured and persistent scene representations (e.g., scene embeddings, map-based anchors, or traffic state descriptors). These enhancements could enable the retrieval of prior user confirmations or preferences in similar contexts, fostering more consistent and personalized ADAS support.

\subsection{Efficiency: Latency and Resource Usage}

\subsubsection{Latency Analysis Across Service Types}

We evaluated the end-to-end reasoning latency across three modular service types:
\begin{itemize}
    \item \textbf{Conversational-Only Service}: Refiner + Responder
    \item \textbf{Conversational ADAS Service}: Refiner + Actuator + Responder
    \item \textbf{Scene-Aware Conversational Service}: Refiner + Vision + Responder
\end{itemize}

For clarity, we refer to the Context Retrieval module (vision-to-text processing) as the Vision module, and the Command Generation module (structured ADAS function calling) as the Actuator module in the following analysis.

All service types share the Refiner and Responder modules, providing a consistent baseline for comparing the impact of the Vision and Actuator modules with minimal confounding factors.

\begin{figure}
    \centering
    \includegraphics[width=0.99\columnwidth]{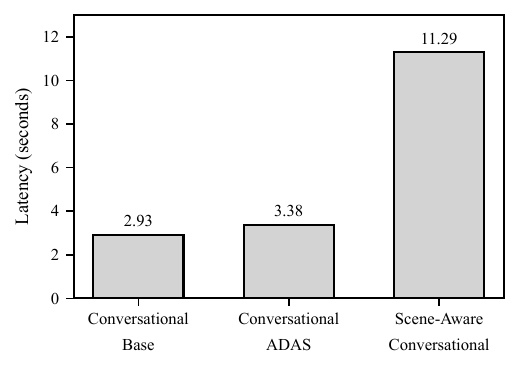}
    \caption{Latency comparison across different service configurations. Scene-aware vision processing introduces significant additional latency, while command generation overhead remains minimal.}
    \label{fig:latency_comparison}
\end{figure}

As shown in Fig.~\ref{fig:latency_comparison}, the Scene-Aware Service exhibited an average latency of approximately 11 seconds, introducing an additional 8 seconds compared to the base Conversational-Only Service (3 seconds).  
The Conversational ADAS Service maintained similar latency to the baseline, indicating that structured function generation incurs minimal additional delay.
The increased latency is primarily attributable to the Vision module's image processing and scene interpretation workload.

\subsubsection{Latency Breakdown per Service}

\begin{figure}
    \centering
    \includegraphics[width=0.99\columnwidth]{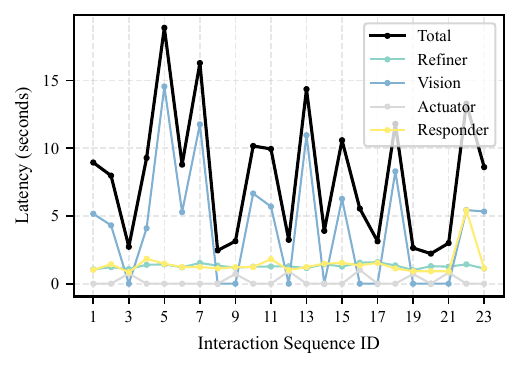}
    \caption{Latency decomposition across interaction sequences, showing the contribution of each module (Refiner, Vision, Actuator, Responder) to the total service processing time. 
    }
    \label{fig:latency_breakdown}
\end{figure}

Fig.~\ref{fig:latency_breakdown} presents per-service latency decomposition across agent modules.  Interaction sequences correspond to sequential conversational turns simulated during driving.
The Refiner and Responder modules consistently contribute 2--5 seconds. The Actuator module adds approximately 1 second.  Vision processing introduces the most variability, with end-to-end latencies ranging from 8 to 17 seconds, particularly in visually complex scenes.

\subsubsection{Token Usage Growth and Context History}

\begin{figure}[t]
    \centering
    \includegraphics[width=0.99\columnwidth]{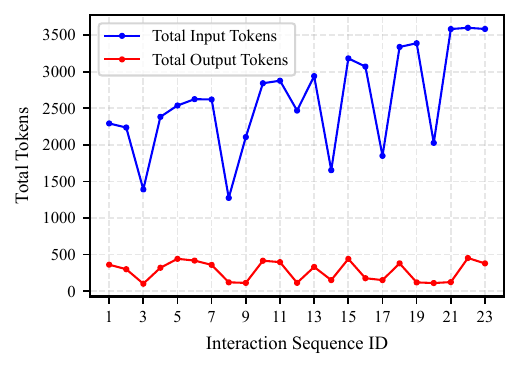}
    \caption{Analysis of total input (uplink) and output (downlink) token usage across interaction sequences, showing input token growth driven by conversation history.}
    \label{fig:token_usage}
\end{figure}

Token usage analysis (Fig.~\ref{fig:token_usage}) shows that total token volume, particularly for input (uplink) tokens, grows over time as dialogue history accumulates.   Output (downlink) tokens remain relatively stable across services, typically under the 500-token maximum configuration.

\subsubsection{Token Distribution Across Agents}

\begin{figure}[t]
    \centering
    \begin{subfigure}[t]{\columnwidth}
        \centering
        \includegraphics[width=\columnwidth]{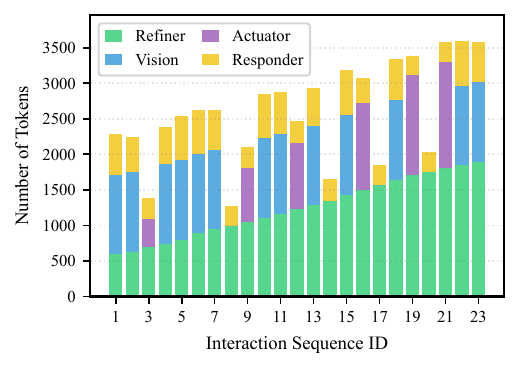}
        \caption{Input tokens per per interaction sequence. The Refiner dominates due to conversation context accumulation.}
        \label{fig:input_tokens}
    \end{subfigure}

    \begin{subfigure}[t]{\columnwidth}
        \centering
        \includegraphics[width=\columnwidth]{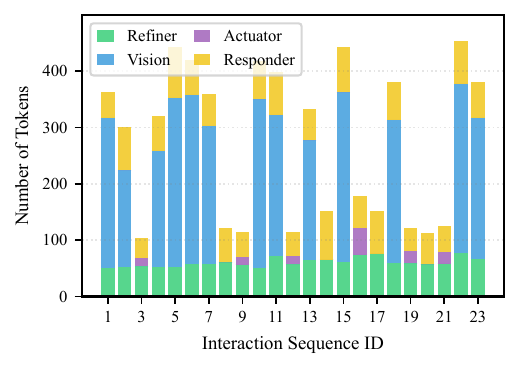}
       \caption{Output tokens per per interaction sequence. The output token volume relatively remained low.  The vision module contributed additional output tokens. 
       }
        \label{fig:output_tokens}
    \end{subfigure}
    \caption{Input and output token distributions across agent modules over 23 service invocations.}
    \label{fig:token_analysis}
\end{figure}

To better understand resource demands per module, we further analyzed input and output token distributions.  
Fig.~\ref{fig:token_analysis} shows the breakdown across 23 complete service invocations.

The Refiner module consistently accounted for the majority of input tokens due to conversation history management.  
The Vision module contributed additional input and output tokens  when scene interpretation was triggered. 
Output token volume remained significantly lower than input volume, with the Vision module contributing the highest peaks during image-grounded interactions. The Refiner and Responder modules consistently generated moderate output tokens across most interactions.

%% file: sections/future_work.tex
\section{Future Work}
\label{sec:future}


Several directions will extend the capabilities of SC-ADAS. First, we plan to deploy lightweight vision and language models on embedded platforms to reduce cloud latency and minimize uplink token transmission. Second, to improve contextual consistency and preference retention, we will integrate Retrieval-Augmented Generation to retrieve past driver interactions and system responses in visually or semantically similar scenes. Finally, we will explore output validation techniques to ensure the safety and reliability of GenAI-generated decisions in real-world applications.

%% file: sections/conclusion.tex
\section{Conclusion}
\label{sec:conclusion}

This paper presented SC-ADAS, a modular framework that integrates Generative AI components, including LLM reasoning, vision-to-text interpretation, and structured function calling to enable real-time, scene-aware driver assistance. The system supports multi-turn dialogue, enabling context-grounded recommendations and ADAS control through natural language.
Evaluation in a simulated environment showed that SC-ADAS effectively interprets diverse road and weather conditions to generate adaptive speed recommendations. The results also reveal trade-offs introduced by scene-aware processing, including increased latency from vision-based context retrieval and uplink token growth due to accumulated dialogue history. Additionally, slight variations in recommendations across revisited scenes suggest the benefit of more structured scene representations to improve consistency.
Overall, SC-ADAS demonstrates the feasibility of combining conversational reasoning, visual context understanding, and modular ADAS control for interpretable and adaptive driver support.


%% file: sections/acknowledge.tex
\section*{Acknowledgment}
The contents of this work only reflect the views of the authors, who are responsible for the facts and the accuracy of the data presented herein. The contents do not necessarily reflect the official views of Toyota Motor North America.
